\documentclass{article}
\usepackage{arxiv}
\usepackage[utf8]{inputenc} % allow utf-8 input
\usepackage[T1]{fontenc}    % use 8-bit T1 fonts
\usepackage{hyperref}       % hyperlinks
\usepackage{url}            % simple URL typesetting
\usepackage{booktabs}       % professional-quality tables
\usepackage{amsfonts}       % blackboard math symbols
\usepackage{nicefrac}       % compact symbols for 1/2, etc.
\usepackage{microtype}      % microtypography
\usepackage{graphicx}
\usepackage[noend]{algpseudocode}
\usepackage{algorithmicx,algorithm}
\usepackage{amsmath}
\usepackage{threeparttable}
\usepackage{cite}
\usepackage{url}
\usepackage{subfig}

\title{Attention augmented differentiable forest\\ for tabular data}
\author{%
Yingshi Chen \\
  Department of Electronic Science\\
  Xiamen University, Xiamen 361005, China \\
  \texttt{gsp@grusoft.com} \\
}

\begin{document}

\maketitle

\begin{abstract}
  Differentiable forest is an ensemble of decision trees with full differentiability. Its simple tree structure is easy to use and explain. With full differentiability, it would be trained in the end-to-end learning framework with gradient-based optimization method. In this paper, we propose tree attention block(TAB) in the framework of differentiable forest. TAB block has two operations, squeeze and regulate. The squeeze operation would extract the characteristic of each tree. The regulate operation would learn nonlinear relations between these trees. So TAB block would learn the importance of each tree and adjust its weight to improve accuracy. Our experiment on large tabular dataset shows attention augmented differentiable forest would get comparable accuracy with gradient boosted decision trees(GBDT), which is the state-of-the-art algorithm for tabular datasets. And on some datasets, our model has higher accuracy than best GBDT libs(LightGBM, Catboost, and XGBoost). Differentiable forest model supports batch training and batch size is much smaller than the size of training set. So on larger data sets, its memory usage is much lower than GBDT model. The source codes are available at https://github.com/closest-git/QuantumForest.
\end{abstract}

\section{Introduction}

Differentiable decision tree\cite{kontschieder2015deep,popov2019neural} brings differentiable properties to classical decision tree. Compared with widely used deep neural networks, tree model still has some advantages. Its structure is very simple, easy to use and explain the decision process. Especially for tabular data, gradient boosted decision trees(GBDT)\cite{friedman2001greedy} models usually have better accuracy than deep networks, which is verified by many Kaggle competitions and real applications. But the classical tree-based models lack of differentiability, which is the key disadvantage compare to the deep neural networks. Now the differentiable trees also have full differentiability. So we could train it with many powerful gradient-based optimization algorithms (SGD, Adam,…). We could use batch training to reduce memory usage greatly. And we could use the end-to-end learning to reduce many preprocess works. Many powerful techniques in the deep learning would also be used in the framework of differentiable trees. For example, we improve the accuracy by the attention mechanism in this paper. We have implemented this method in the open-source library QuantumForest. Experiments on large datasets verified its effectiveness. 

\section{RELATED WORK}
In recent years, different research teams\cite{kontschieder2015deep,popov2019neural,tanno2018adaptive,yang2018deep,lay2018random,feng2018multi,silva2019optimization,hazimeh2020tree} have proposed different models and algorithms to implement the differentiability. Compared with the previous work, the main contribution of our work is to improve differentiable forest model with tree-based attention mechanism.

\subsection{Differentiable forest}
\cite{kontschieder2015deep} is a pioneering work in the differentiable decision tree model. They present a stochastic routing algorithm to learn the split parameters via backpropagation. The tree-enhanced network gets higher accuracy than GoogleNet \cite{szegedy2015going}, which was the best model at the time of its publication. \cite{popov2019neural} introduces neural oblivious decision ensembles (NODE) in the framework of deep learning. The core unit of NODE is oblivious decision tree, which uses the same feature and decision threshold in all internal nodes of the same depth. There is no such limitation in our algorithm. As described in section \ref{sec:algorithm}, each internal node in our model could have independent feature and decision threshold. \cite{tanno2018adaptive} presents adaptive neural trees (ANTs) constructed on three differentiable modules(routers, transformers and solvers). In the training process, ANTs would adaptively grow the tree structure. Their experiment showed its competitive performance on some testing datasets. \cite{yang2018deep} presents a network-based tree model(DNDT). Their core module is soft binning function, which is a differentiable approximation of classical hard binning function. \cite{lay2018random} propose random hinge forests or random ferns with differentiable ReLU-like indicator function. Then the loss function would be optimized end-to-end with stochastic gradient descent algorithm. \cite{feng2018multi} propose stacked GBDT layers(mGBDTs). Their model can be jointly trained by a variant of target propagation across layers, without the need to derive backpropagation nor differentiability. \cite{silva2019optimization} presents optimization method to get policy gradient of decision tree in the framework of reinforcement learning. Their tree models trained with policy gradients would get comparable and even superior performance against MLP baselines. In some sense, soft decision tree \cite{irsoy2012soft} can be regarded as the earliest prototype of a differential tree, which has only one differentiable gating function. 

\subsection{Gradient boosted decision trees}
Gradient boosted decision trees(GBDT)\cite{friedman2001greedy} is the state-of-the-art algorithm for tabular datasets, which has been verified in many real applications. In most Kaggle competitions on tabula data, GBDT based solution always has higher accuracy than any other solutions. In general, the best performing GBDT libraries are lightGBM\cite{ke2017lightgbm}, Catboost\cite{dorogush2018catboost} and XGBoost\cite{chen2016xgboost}. The GBDT algorithm would learn a series of weak learners to predict the output. The weak-learner in GBDT is standard decision tree, which is lack of differentiability. The significant drawback of tree-based approaches is that they usually do not allow end-to-end optimization and employ greedy, local optimization procedures for tree construction. Thus, they cannot be used as a component for pipelines, trained in an end-to-end fashion.

\subsection{Feature selection, routing function and attention mechanism}
There are usually three steps in the routing(split) function. First, some features are selected, then calculate a value through some transformation, and finally compared with the threshold. Many different algorithms have been proposed in the previous implementation. The feature transformation in \cite{kontschieder2015deep} is a fully connected layer. Each routing function corresponds to an output neuron node of FC layer. The weights of FC layer are dense and there are no special sparse transformation or attention mechanism.\cite{popov2019neural} use entmax transformation \cite{peters2019sparse}. In their experiment, entmax would make feature weight matrix much sparser than classical softmax or sparsemax transformation \cite{martins2016from}. The feature transformation in \cite{tanno2018adaptive} is small CNN. So all the features are used in the routing process. And there is no attention mechanism in their implementation. \cite{yang2018deep,lay2018random,feng2018multi,silva2019optimization} all use only one feature to make decision. No feature transformation or attention mechanism in their implementation. 

\subsection{Testing dataset}
The data sets tested in previous papers are quite different. Some papers \cite{kontschieder2015deep,tanno2018adaptive} focus on image classification. So their testing datasets are image datasets: CIFAR, ImageNet, MNIST... For the tabular dataset, Some research\cite{lay2018random,feng2018multi} only tested with some small datasets: Titanic, Iris, Pima, abalone, Income, Protein... That is, the samples in the training sets are all less than 100,000. \cite{popov2019neural,yang2018deep} has used large testing datasets. In this paper, we use six large open-source tabular datasets: Epsilon, YearPrediction, Higgs, Microsoft, Yahoo, Click. All train sets have more than 400,000 samples.

\section{Differentiable forest with tree attention block}\label{sec:algorithm}
\begin{figure}[H]
	\centering	
	\subfloat[Simplest differentiable tree with only three nodes (one root node with two child nodes)]{{\includegraphics[width=5cm]{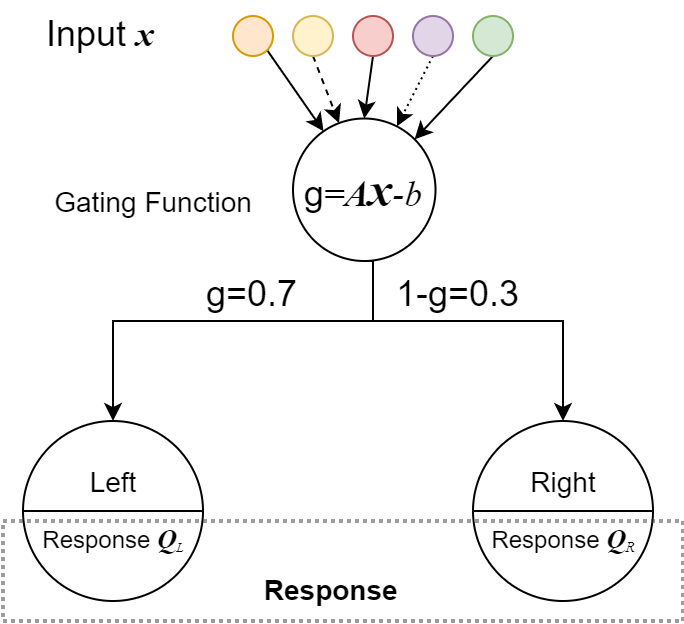} }}
  \qquad
  \subfloat[Differentiable tree and its probability at each leaf nodes. In this sample, The input $x$ would reach $n_3$ with probability $1-g_1$ and reach $n_6$ with probability $(1-g_1) g_3$]{{\includegraphics[width=5cm]{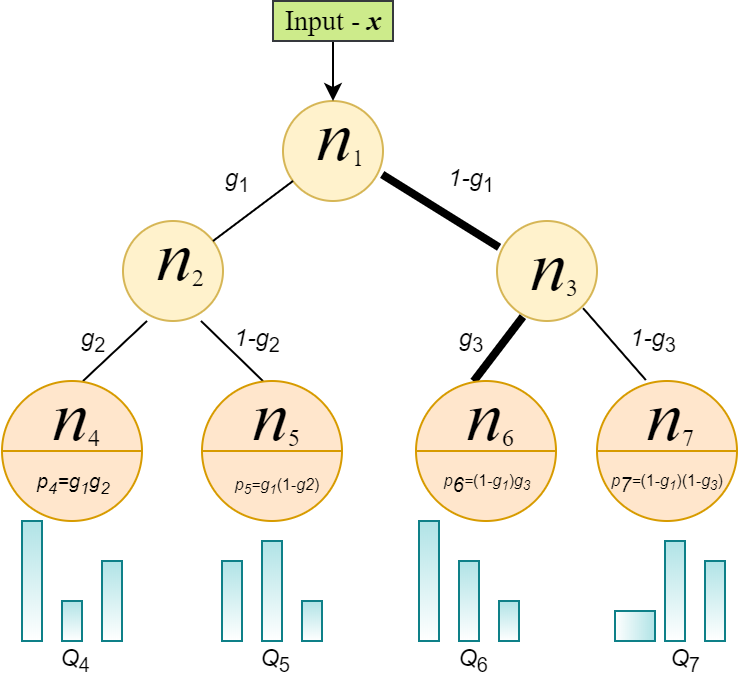} }}
  \caption{Differentiable tree}
  \label{fig:trees}
\end{figure}

To describe the problem more concisely, we give the following formulas and symbols:

For a dataset with N samples  $\boldsymbol{X}= \left \{ x \right \}$  and its target $ \boldsymbol{Y}= \left \{ y \right \} $. Each $ x $ has M attributes, $  x=\left [ x_{1},x_{2},\cdots,x_{M} \right ]^{T} $. The differentiable forest model would learn $K$ differentiable decision trees $  \left [ T_{1},T_{2},\cdots,T_{K} \right ] $  to minimize the loss between the target $y$  and prediction $\hat{y}$.
\begin{equation}
\hat{y}=\frac{1}{K}\sum_{h=1}^{K}T^h\left ( x \right )
\end{equation}
Figure \ref{fig:trees}.(a) shows the simplest case of differentiable tree. It has only one gating function $ g $ to test the input $ x $. There are three nodes. The root node represented the gating function with some parameters. It has two child nodes, which are represented as $\left\lbrace left,right \right\rbrace$. The root node would redirect input $ x $ to both $\left\lbrace left,right \right\rbrace$ with probabilities calculated by the gating function $g$. Formula \ref{F:gate} gives the general definition of gating function, where $A\in R^{M}$  is a learnable weight parameter for each attribute of $ x, b $ is a learnable threshold. $\sigma$ would map $ Ax-b $ to probability between [0,1], for example, the sigmoid function.
\begin{equation}\label{F:gate}
g\left ( A,x,b \right )=\sigma \left ( Ax-b \right )
\end{equation}
So as shown in Figure \ref{fig:trees}.(b), The sample $x$  would be directed to each nodal $ j $ with probability $ p_j $. And finally, the input x would reach all leaves. For a tree with depth $ d $, we represent the path as $ {n_1,n_2,\cdots,n_d } $, where $ n_1 $ is the root node and $ n_d $  is the leaf node $ j $. $ p_j $ is just the product of the probabilities of all nodes in this path:
\begin{equation}\label{F:p}
  p_{j}=\prod_{n\in \left \{ n_{1},\cdots ,n_{d} \right \}}^{}g_{n}
\end{equation}
It's the key difference with the classic decision trees, in which the gating function $g$ is just the heave-side function, so each $x$ always get only one state, either left or right. And finally, the input $x$ would only reach one leaf. 

The output at leaf node $j$ is represented by reponse vector $\boldsymbol{q}_j=\left ( q_{j_1},q_{j_2}, \cdots,q_{j_F} \right )^T$ . Then the output of tree $h$ is just the probability average of these responses. 
\begin{equation}\label{F:leaf_y}
Q_h\left ( x \right )=\sum_{j\ is\ leaf\ of\ h}p_j\boldsymbol{q}_j
\end{equation}

A single tree is a very weak learner, so we should merge many trees to get higher accuracy, just like the random forest or other ensemble learning method. The final prediction $y$ is weighted summary of all trees. In the simplest case, the weight is always $1/K$, $\hat{y}$ is just the average result.
\begin{equation}\label{F:Y}
  \hat{y}\left ( x \right )=\frac{1}{K}\sum_{h=1}^{K}Q_h\left ( x \right )
\end{equation}

In our model, we tree attention block(TAB) to learn the weight. As formula \ref{F:wY} shows, the weight $\omega$ of each tree $h$ is the output of TAB block. The detailed algorithm is given in the subsection \ref{subsec:attention-tree}.
\begin{equation}\label{F:wY}
\begin{split}
\hat{y}\left ( x \right )&=\frac{1}{K}\sum_{h=1}^{K}\omega Q_h\left ( x \right )\\
\omega&=regulate(squeeze(\left [ Q_1,Q_2,\cdots ,Q_K \right ]))
\end{split}
\end{equation}

Let $\Theta$ represents all parameters $(A,b,Q)$, then the final loss would be represented by the general form of formula \ref{F:loss}

\begin{equation}\label{F:loss}
  L(\Theta:x,y)=\frac{1}{K}\sum_{h=1}^{K}L_h(\Theta:x,y)=\frac{1}{K}\sum_{h=1}^{K}L_h(A,b,Q:x,y)
\end{equation}

where $L:\mathbb{R}^F \mapsto \mathbb{R}$ is a funciton that maps vector to object value. In the case of classification problem, the classical function of $L$ is cross-entropy. For regression problem, $L$ maybe mse, mae, huber loss or others. To minimize the loss in formula \ref{F:loss}, we use stochastic gradient descent(SGD) method to train this model. 

\subsection{Tree attention block}\label{subsec:attention-tree}
Our attention mechanism is inspired by the channel attention mechanism in the famous SENet \cite{hu2018squeeze,wang2019eca,woo2018cbam}. Although the structure and algorithm of differentiable forest and convolutional neural network(CNN) are quite different. There are some similarity. Each channel(tree) represents some features of samples. Since all these features constitute a complete description, there must be deep relations between these channels(trees). Attention mechanism is an efficient and lightweight technique to discover the internal connections in these channels(trees).

\subsubsection{Channel attention block in SENet}
SENet uses squeeze-and-excitation(SE) blocks to implement channel attention mechanism. The classical deep convolutional neural networks is composed of many convolution blocks. Let the output of one block is $\mathbf{U}\in \mathbb{R}^{H\times W\times C}$, that is, $C$ channels with spatial dimensions $H\times W$. As formula \ref{F:se_net} shows, SE block has two operations, squeeze and excitation. The squeeze function is actually global average pooling, which would extract the characteristic of each channel. The excitation function would learn nonlinear interaction of channels by two FC layers and relu layer.
\begin{equation}\label{F:se_net}
  \begin{split}
  \omega&=excitation(squeeze(\mathbf{U}))\\
  squeeze(\mathbf{U})&=\left [ \frac{1}{HW}\sum_{i=1,j=1}^{W,H} U_{i,j}^{c} \quad c=0,1,\cdots ,C\right ]^{T}\\
  excitation\left ( \mathbf{z} \right )&=\sigma\left ( \mathbf{W}_2\delta \left ( \mathbf{W}_1\left (\mathbf{z}  \right ) \right ) \right )
  \end{split}
\end{equation}
where $\sigma$ is sigmoid function, $\delta$ is relu function \cite{nair2010rectified}, $\mathbf{W}_1,\mathbf{W}_2$ is the parameters of two FC layers.

\subsubsection{Tree attention block(TAB)}
As illustrated in Figure 2, we use tree-based attention block to model the relation of all trees. Let the response vector  of each tree is $\hat{y}_h=\left ( q_{h_1},q_{h_2}, \cdots,q_{h_F} \right )$, the output of all $K$ trees is $\hat{\mathbf{Y}}=\left [ \hat{y}_1,\hat{y}_2,\cdots ,\hat{y}_K \right ]^{T}$. In the squeeze operator, we just use the average of each response as tree descriptor. This would transform $\hat{\mathbf{Y}}$ to a descriptor vector with size $1 \times K$. The regulate operator has two fully connected layers, with a non-linear relu layer in the middle. This would map the descriptor vector to a weight vector $\left [ \omega_1, \omega_2,\cdots , \omega_K \right ]$. We then apply this weight vector to $\hat{\mathbf{Y}}$ to get final output $\left [ \omega_1\hat{y}_1,\omega_2\hat{y}_2,\cdots ,\omega_K\hat{y}_K \right ]^{T}$.    
\begin{figure}[H]
	\centering	
	\includegraphics[width=4in]{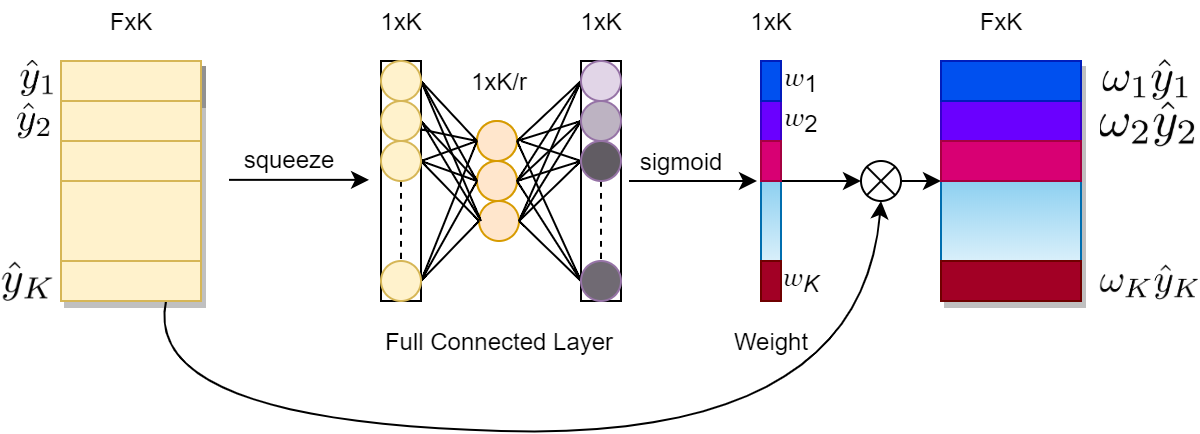}
	\caption{Tree attention block for differentiable forest}
\end{figure}

Formula \ref{F:se_tree} lists the detailed process of tree attention block. Compare formula \ref{F:se_net} and formula \ref{F:se_tree}, the key difference is the input of squeeze function.
\begin{equation}\label{F:se_tree}
  \begin{split}
  \omega&=regulate(squeeze(\hat{\mathbf{Y}}))\\
  squeeze(\hat{\mathbf{Y}})&=\left [ mean(\hat{y}_h) \quad h=0,1,\cdots ,K\right ]^{T}\\
  regulate\left ( \mathbf{z} \right )&=\sigma\left ( \mathbf{W}_2\delta \left ( \mathbf{W}_1\left (\mathbf{z}  \right ) \right ) \right )
  \end{split}
\end{equation}
where $\sigma$ is sigmoid function or softmax function, $\delta$ is relu function \cite{nair2010rectified}, $\mathbf{W}_1\in \mathbb{R}^{K/r\times K}$ $\mathbf{W}_2\in \mathbb{R}^{K\times K/r}$ is the parameters of two FC layers($r$ is the reduction ratio).

\subsection{Learning algorithm}
Based on the general loss function defined in formula \ref{F:learn_batch}, we use stochastic gradient descent method \cite{kingma2014adam,ma2018quasi} to reduce the loss. As formula \ref{F:learn_batch} shows, update all parameters $\Theta$ batch by batch:
\begin{equation}\label{F:learn_batch}
\Theta^{t+1}= \Theta^t - \eta\frac{\partial L }{\partial \Theta}\sum_{\left (x,y  \right )\in \mathfrak{B} }\frac{\partial L }{\partial \Theta}\left ( \Theta^t;x,y \right )
\end{equation}
where $ \mathfrak{B} $ is the current batch, $ \eta $ is the learning rate, $ (x,y) $ is the sample in current batch.

This is similar to the training process of deep learning. Some hyperparameters (batch size, learning rate, weight decay,drop out ratio…) need to be set. All the training skills from deep learning could be used. For example, the batch normalization technique, drop out layer. We find QHAdam \cite{ma2018quasi} would get a few higher accuracy than Adam\cite{kingma2014adam}. So QHAdam is the default optimization algorithm in QuantumForest.

\begin{algorithm}[H]
  \caption{Learning algorithm of differentiable forest}
  \hspace*{0.02in} {\bf Input:}
  input training, validation and test dataset \\
  \hspace*{0.02in} {\bf Output:} 
  learned model
  \begin{algorithmic}[1]
  \State Init feature weight mattrix $A$
  \State Init response $Q$ at each leaf nodes
  \State Init threshhold values $b$
  \State Init the paramerters in attention block: $\mathbf{W}_1$,$\mathbf{W}_2$

  \While{not converge}
    \For{each batch}
        \State Calculate gating value at each internal node
        \State \quad $g\left ( A,x,b \right )=\sigma \left ( Ax-b \right )$
        \State Calculate probability at each leaf node
        \State \quad $p_{j}=\prod_{n\in \left \{ n_{1},\cdots ,n_{d} \right \}}^{}g_{n}$
        \State Get response of each tree
        \State \quad $\hat{y}_h\left ( x \right )=\sum_{j\ is\ leaf}p_jQ_j$
        \State Get the weight of each tree from attention mechanism
        \State \quad $\omega=\sigma\left ( \mathbf{W}_2\delta \left ( \mathbf{W}_1\left (squeeze(\left [ \hat{y}_1,\hat{y}_2,\cdots ,\hat{y}_K \right ])  \right ) \right ) \right )$
        \State Get the output of current batch
        \State \quad $\hat{y}\left ( x \right )=\frac{1}{K}\sum_{h=1}^{K}\omega \hat{y}_h\left ( x \right )$
        \State Calculate the loss $L$
        \State Backpropagate to get the gradient
        \State \quad $\left ( \delta A,\delta b,\delta Q,\delta \mathbf{W}_1,\delta \mathbf{W}_2 \right )\Leftarrow \delta L$
        \State Update the parameters: $A$, $Q$, $b$, $\mathbf{W}_1$, $\mathbf{W}_2$
    \EndFor
    \State Evalue loss at validation dataset
  \EndWhile
  \State
  \Return learned model
  \end{algorithmic}
  \end{algorithm}

\section{Results and discussion}

To verify our model and algorithm, we test its performance on six large datasets. Table \ref{tab:datasets} lists the detail information of these datasets. We split each dataset into training,validation and test sets. The training/validation set is used to learn differentialble forest models. The test set is used to evaluate the performance of the learned models.

\begin{table}[H]
  \centering
  \caption{Six large tabular datasets}
  \label{tab:datasets}
  \resizebox{\textwidth}{!}{%
  \begin{tabular}{|r|c|c|c|c|c|c|}
  \hline
              & Higgs \cite{baldi2014searching}     & Click \cite{Click}     & YearPrediction\cite{Year}  & Microsoft \cite{qin2013introducing} & Yahoo \cite{Yahoo}          & EPSILON \cite{epsilon}              \\ \hline
  Training      & 8.4M & 800K & 309K & 580K & 473K & 320K \\ \hline
  validation & 2.1M & 100K & 103K & 143K & 71K  & 80K  \\ \hline
  Test       & 500K & 100K & 103K & 241K & 165K & 100K \\ \hline
  Features   & 28   & 11   & 90   & 136  & 699  & 2000 \\ \hline
  Problem     & Classification & Classification & Regression           & Regression     & Regression          & Classification        \\ \hline
  Description & UCI ML Higgs   & 2012 KDD Cup   & Million Song Dataset & MSLR-WEB 10k   & Yahoo LETOR dataset & PASCAL Challenge 2008 \\ \hline
  \end{tabular}%
  }
  \end{table}

\subsection{Accuracy}
We compare the accuracy of QuantumForest with the following libraries:

1)	Catboost\cite{dorogush2018catboost}. A GBDT library which uses oblivious decision trees as weak learners. We use the open-source implementation at https://github.com/catboost/catboost.
\\2)	 XGBoost\cite{chen2016xgboost}. We use the open-source implementation at https://github.com/dmlc/xgboost
\\3)	NODE\cite{popov2019neural}. A new neural oblivious decision ensembles for deep learning. We use the open-source implementation at https://github.com/Qwicen/node
\\4)	mGBDT\cite{feng2018multi}: Multi-layered gradient boosting decision trees by [21]. We use the open-source implementation at https://github.com/kingfengji/mGBDT
\\5)	LightGBM\cite{ke2017lightgbm}: A fast, distributed, high performance gradient boosting framework. We use the open-source implementation at https://github.com/Microsoft/LightGBM

LightGBM, Catboost and XGBoost are the best GBDT libs, which are the state-of-the-art tools for the tabular datasets. NODE\cite{popov2019neural} is based on the  differentiable oblivious forest. In some sense, the model of NODE is a special version of our model. That is, the nodes in each layer share only one gating function. 

\begin{table}[H]
  \centering
  \caption{Accuracy comparison*}
  \label{tab:accuracy}
  \resizebox{\textwidth}{!}{%
  \begin{tabular}{|r|c|c|c|c|c|c|}
  \hline
           & Higgs  & Click           & YearPrediction & Microsoft & Yahoo  & EPSILON \\ \hline
  CatBoost & 0.2434 & 0.3438          & 80.68          & 0.5587    & 0.5781 & 0.1119  \\ \hline
  XGBoost  & 0.2600 & 0.3461          & 81.11          & 0.5637    & 0.5756 & 0.1144  \\ \hline
  LightGBM      & \textbf{0.2291} & 0.3322          & 76.25          & 0.5587           & \textbf{0.5576} & 0.1160          \\ \hline
  NODE     & 0.2412 & \textbf{0.3309} & 77.43          & 0.5584    & 0.5666 & \textbf{0.1043}  \\ \hline
  mGBDT    & OOM    & OOM             & 80.67          & OOM       & OOM    & OOM     \\ \hline
  QuantumForest & 0.2467          & \textbf{0.3309} & \textbf{74.02} & \textbf{0.5568} & 0.5656          & 0.1048 \\ \hline
  \end{tabular}%
  }
  \begin{tablenotes}
    \small
    \item *Some results are copied form the testing results of NODE \cite{popov2019neural}.
  \end{tablenotes}
  \end{table}

Table \ref{tab:accuracy} listed the accuracy of all libraries. All libraries use default parameters. For each dataset, QuantumForest uses 1024 trees, the batch size is 512, the default learning rate is 0.002. 

In general, some libraries perform better on certain data sets, while others perform better on others. LightGBM is the winner of 'Higgs' and 'Yahoo' datasets. NODE is the winner of 'Click' datasets. Our model performs best on the 'Click', 'YearPrediction','Microsoft', and 'EPSILON' datasets.  mGBDT always failed because out of memory(OOM) for most large datasets. Both NODE and QuantumForest have higher accuracy than CatBoost and XGBoost. It is a clear sign that differentiable forest model has more potential than classical GBDT models. 

The differentiable forest model has only been developed for a few years and is still in its early stages. We are sure its performance would increate a lot. That doesn't mean it would be best in all cases. As the famous no free lunch theorem, some lib would perform better in some datasets and maybe poor in other datasets. Anyway, QuantumForest shows the great potential of differentiable forest model. 

\subsection{Memory Usage}
In this subsection, we compare the memory usage between differentiable forest model (QuantumForest) and three GBDT libraries(CatBoost, XGBoost, lightGBM). Table \ref{tab:memory} listed the memory used by these libraries at six datasets. In all cases, XGBoost requires the most memory. In smaller datasets, lightGBM/CatBoost needs less memory. In bigger datasets('EPSILON' and 'Higgs'), QuantumForest needs much less memory than GBDT libraries. This reflects the essential difference between these two models. QuantumForest uses batch training. The number of samples per batch is fixed. The main factors for memory usage are feature numbers and trees number. On the other hand, GBDT models always try to load all training samples into memory. The main factors for memory usage are the size of training set(sample numbers, feature numbers). All GBDT library has tow sub-sampling parameters to reduce the number of samples and features in the training process. But the ratio of sub-sampling cannot take an arbitrarily small value. For large datasets, the size of sub-samples would be much larger than the batch size in QuantumForest. And QuantumForest also supports feature subsample technique to reduce memory usage. For the differentiable forest model, no matter big data or small data set, the batch size is all 512, the tree number is all 1024. So for smaller datasets, the memory usage is higher than GBDT models. For bigger datasets, batch training based differentiable forest model would use less memory than GBDT model. 

% \usepackage{graphicx}
\begin{table}[H]
  \centering
  \caption{Comparison of memory usage on 6 datasets (MB)}
  \label{tab:memory}
  \resizebox{\textwidth}{!}{%
  \begin{tabular}{|c|c|c|c|c|c|c|}
  \hline
                & Higgs         & Click        & YearPrediction & Microsoft     & Yahoo         & EPSILON       \\ \hline
  CatBoost      & 5234          & \textbf{388} & 985            & 1282          & 3268          & 11481         \\ \hline
  XGBoost       & 8860          & 590          & 1435           & 3376          & 9555          & 22958         \\ \hline
  lightGBM      & 4503          & 399          & \textbf{905}   & \textbf{1244} & \textbf{3093} & 14336         \\ \hline
  QuantumForest & \textbf{3857} & 2660         & 2971           & 3118          & 4060          & \textbf{7024} \\ \hline
  \end{tabular}%
  }
  \end{table}

\begin{figure}[H]
	\centering	
	\includegraphics[width=5in]{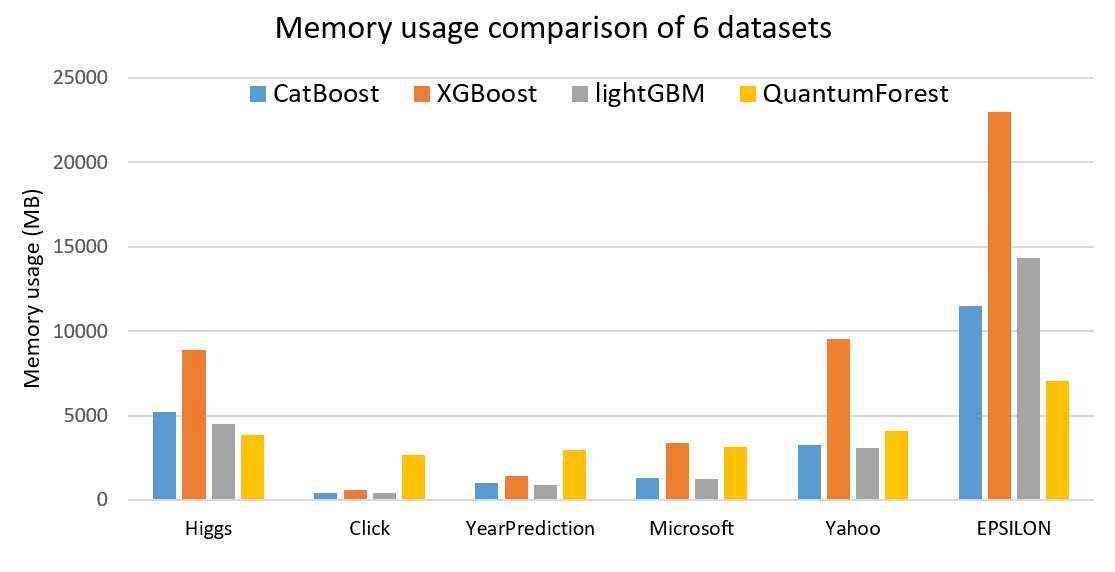}
	\caption{Memory usage at six datasets}
\end{figure}

\section{Conclusion}
In this paper, we propose tree-based attention mechanism in the framework of differentiable forest. The tree attention block(TAB) would learn the importance of each tree to improve accuracy. Our experiment on large testing data verified its effectiveness. Since TAB is a lightweight and general module, it can be integrated into any tree-based architectures(For example, random forest) with little extra overheads. We hope this would become an important component of various tree-based models. To further study and improve this algorithm, we developed an open-source package QuantumForest. The codes are available at https://github.com/closest-git/QuantumForest.

\bibliographystyle{ieeetr}
% \bibliography{references} % 导入lib，ref为“ref.lib"的文件名

\end{document}